# Reading Miscue Detection in Primary School through Automatic Speech Recognition

*Lingyun Gao*[1], *Cristian Tejedor-Garcia*[1], *Helmer Strik*[1], *Catia Cucchiarini*[1]

[1]Centre for Language Studies, Radboud University Nijmegen, Netherlands

{lingyun.gao, cristian.tejedorgarcia, helmer.strik, catia.cucchiarini}@ru.nl

## Abstract

Automatic reading diagnosis systems can benefit both teachers for more efficient scoring of reading exercises and students for accessing reading exercises with feedback more easily. However, there are limited studies on Automatic Speech Recognition (ASR) for child speech in languages other than English, and limited research on ASR-based reading diagnosis systems. This study investigates how efficiently state-of-the-art (SOTA) pretrained ASR models recognize Dutch native children speech and manage to detect reading miscues. We found that Hubert Large finetuned on Dutch speech achieves SOTA phoneme-level child speech recognition (PER at 23.1%), while Whisper (Faster Whisper Large-v2) achieves SOTA word-level performance (WER at 9.8%). Our findings suggest that Wav2Vec2 Large and Whisper are the two best ASR models for reading miscue detection. Specifically, Wav2Vec2 Large shows the highest recall at 0.83, whereas Whisper exhibits the highest precision at 0.52 and an F1 score of 0.52.

## 1. Introduction

Recent advances in Automatic Speech Recognition (ASR) have brought many potential applications that seemed appealing, but that appeared to be very complex, within reach [1]. One such case is using ASR in primary school reading education [2], which started almost twenty years [3, 4], but faced many problems with child speech recognition [5, 6] and miscue detection [7, 8]. However, growing concerns about decreasing levels of reading proficiency [9, 10, 11], emphasize the need for innovative solutions to boost reading instruction.

State-of-the art (SOTA) ASR technology may provide interesting solutions [12] for practice and remedial teaching [13]. In this paper, we investigate the usability of ASR technology for detecting reading miscues [14] in read aloud speech of pupils in Dutch primary schools. As existing datasets are insufficient for training specific children ASR models based on SOTA ASR architectures, we apply pretrained foundation ASR models as core modules for downstream tasks, which is a common approach [15]. While many different ASR models show excellent recognition performance on adult speech, limited research exists on their performance with child speech [16, 17, 18, 19, 20]. The research questions we address in this paper are:

RQ1: *How effectively do pretrained ASR models, initially trained on adult speech, recognize speech of Dutch native children in primary school?*

RQ2: *To what extent do pretrained ASR-based systems for Dutch demonstrate efficacy in reading miscue detection tasks within primary school settings?*

We answer our two research questions by conducting a two-phase study. In the first phase, we evaluate pretrained SOTA ASR models for recognizing children read speech to investigate their potential for detection tasks. In the second phase, we investigate ASR-based reading miscue detection in two steps: detecting general reading errors (substitution, insertion, deletion of pronunciation) and then identifying reading miscues, which includes a range of specific substitution and insertion errors [14]. In particular, insertions in reading miscues form a subset of the insertions in reading errors. If insertion errors are made where a child correctly reads a word in subsequent attempts following initial restarts or repetitions, they are not categorized as insertion miscues. This practice aligns with the conventions observed in administering Dutch reading tests [21].

## 2. Methodology

### 2.1. Evaluation Dataset and Data Preprocessing

This paper employs reading speech data from Dutch native primary school pupils from the Jasmin-CGN Corpus [22] as the evaluation dataset. This contains recordings of children reading aloud at their mastery reading level, which were aligned with manual orthographic and phonemic annotations, from 71 primary school pupils (age range = 6-13 years old), consisting of 35 female and 36 male children. We employ the prompt text, the reading miscue and reading strategy annotations for the first read story (data access can be found in [14]). We trim the recordings to contain only the first story read by children.

In total, the dataset comprises 2.05 hours of speech with 14,251 reading attempts. 11,322 attempts are labeled as correctly read words, while 615 reading attempts are labeled as incorrect reading words and part-of-words, along with attempts indicating partially read correctly words and inaudible words.

### 2.2. ASR Models, Metrics and Tools

We assess the effectiveness of pretrained SOTA Dutch ASR models in recognizing native children speech at both the word and phoneme levels. We employ selected ASR models to predict phoneme/word-level transcriptions, coupled with the Speech Recognition Toolkit SCTK [23]. Phoneme error rate (PER) and word error rate (WER) are utilized as evaluation metrics to assess recognition performance on phoneme-level and word-level, respectively.

Additionally, we use SCTK to align the prompt text with the word transcription in the Jasmin dataset (prompt-Jasmin alignment) to count the number of reading errors made by the children. Similarly, we employ prompt-ASR alignment to count the number of reading errors predicted by the ASR. We use the Er-ror Ratio, calculated as the number of predicted errors divided by the number of true errors, to assess the discrepancy between ASR-predicted errors and actual errors.

We conduct normalization procedures on transcriptions before computing WER, PER and Error Ratio. For phoneme transcriptions, the normalization steps include removing non-verbal cues in the Jasmin transcriptions, and mapping the ASR transcriptions using the IPA alphabet to the CGN alphabet used for the Jasmin transcriptions. For the prompt and ASR/Jasmin word transcriptions, we remove punctuation and capitalization, and standardize Dutch words/numericals with different formats.

We evaluate the three best-performing open-source phoneme-level ASR models for Dutch in the literature as of January 2024: Wav2Vec2 Base[16], Wav2Vec2-XLSR Large [24] and Hubert Large[18]. The first two models are pretrained on Dutch speech, accessible via Hugging Face [24, 25]. We finetuned the English model Hubert Large on Dutch speech using Common Voice Corpus 16 [26] following the scripts in [25] since the authors did not release the model publicly after reporting their results. At the word level, we assess the performance of the top four publicly available ASR models for Dutch as of January 2024: Wav2Vec2 Large [27] and XLSR Large [28] finetuned both with Dutch speech, MMS with a Dutch adapter [17], and Whisper (Faster Whisper Large-v2) [20].

The computing infrastructure comprises one Nvidia Tesla V100 16GB running Ubuntu 18.04 and utilizing Hugging Face's pipeline with a chunk length set to 10 seconds. The average inference runtime is 1 hour for MMS and 0.5 hours for each of the other models.

### 2.3. Reading Miscue Detection

Detection of word-level reading miscue in Table 1 is investigated. In the first step, we explore general reading error detection. Using prompt-ASR alignment, we extract predicted error pairs: Error type (substitution, deletion, and insertion) and predicted error locations relative to the prompt. Similarly, we extract ground-truth error pairs from prompt-Jasmin alignment. We adopt the loose criterion for error detection following [29], wherein a reading error is considered detected if both the error type and error location match in the prediction and ground-truth.

We evaluate error detection performance, including precision, recall, and F1 score, for both overall and within each error category, using four ASR models employed in word-level speech recognition.

For reading miscue detection, we perform miscue analysis on detected general errors in Jasmin transcriptions and in ASR output and categorize them to miscue in Table 1. Specifically, for a detected reading error, we compare word in prompt and word in transcription by (1) Computing string cosine similarity as orthographic similarity score. If the score exceeds the threshold at 0.8, the error is tagged as OS. (2) Similarly, calculating semantic similarity score using pretrained Word2Vec2 similarity [30] with threshold at 0.7. (3) Testing insertions as substring within 5 subsequent words in prompt for restarts or new words. After obtaining miscue labels, we test miscue detection similarly to previous general error detection.

To gain insights into ASR performance, words in the Jasmin transcription are categorized as correctly read words, restarts (part-of-words), incorrectly read and others, utilizing annotations in the reading error tier and reading strategy. We locate those words to find the predictions from the alignment between ASR and ground-truth, and then recognition accuracy is calculated for the first three abovementioned categories.

Table 1: *Selected word-level reading miscue types for our research, which is a subset of miscue types reported by [14]*

| Reading Miscue Type |
| --- |
| Substitute a word in prompt by another existing dutch word which was semantically identical (**SS**) |
| Substitute a word in prompt by another existing dutch word which was orthographically similar (**OS**) |
| Replace a word in prompt by another existing dutch word which was not orthographically or semantically similar (**O**) |
| Insertion of extra word not in the prompt ($I_m$) |
| Deletion of word: a word in the prompt is not read (**D**) |

## 3. Results

### 3.1. Phoneme-Level Analysis

A performance comparison at the phoneme level of the output of the three phoneme-ASR models is reported for children Dutch speech (Jasmin corpus) and adult (Common Voice Dutch test set) in Table 2. In line with the literature [5, 6], results show that adult speech recognition presents lower PER than children speech recognition in all cases. The finetuned Hubert Large model excels in both child (23.1%) and adult (6.3%) phoneme-level recognition.

Table 2: *ASR performance comparison at the phoneme-level*

| Model | PER % (Child) | PER % (Adult) |
| --- | --- | --- |
| Wav2Vec2 Base | 32.6 | 20.3 [25] |
| XLSR Large | 36.1 | 19.6 [24] |
| Hubert Large | **23.1** | **6.3** |

To understand the performance of the Hubert Large model, we conduct an analysis at the phoneme-level in Table 3. We present a comparison of the 10-most frequent substitution (confusion pairs), deletion, and insertion errors when recognizing child and adult speech. Gray cells are employed to emphasize errors present in the top-10 frequent errors by children, but which are absent in the top-10 frequent errors of adults. Confusion pairs *x->G* means /x/ is accepted by the model to replace /G/. Overall, the top-10 most frequent substitution pairs, deletion errors, and insertion errors exhibit considerable similarity between adult and child. The largest relatively decreases in recognition accuracy are /**G**/ (0.74), /**S**/ (0.66), /**z**/ (0.38), /**v**/ (0.30), /**Z**/ (0.16), /**h**/ (0.13), /**N**/ (0.12), /**t**/ (0.11) and /**@**/ (0.09). Notably, vowels and consonants, particularly /**G**/, /**z**/, and /**v**/, seem particularly challenging to recognize accurately in child speech.

### 3.2. Word-Level Analysis

A comparison at the word level of the output of the four test word-level ASR models with children (Jasmin corpus) and adult (Common Voice Dutch testset) Dutch speech is reported on Table 4. Adult WER strongly correlates with child WER, with only a small absolute decrease observed from adult to child. The Whispermodel excels in both child (9.8%) and adult (6.7%)) word-level recognition. In the last column, Error Ratio represents the number of predicted reading errors by each model divided vs number of ground truth reading errors. For the three Wav2Vec2-family models, Error Ratio exceeds 1, indicating more errors detected, while Whisper's ratio at 0.58 indicates fewer errors detected.

Table 3: *Top-10 recognition errors at phoneme-level*

| Confusion pairs(C): Cnum | | | | Deletion(D):Dnum | | | | Insertion(I):Inum | | | |
|---|---|---|---|---|---|---|---|---|---|---|---|
| child | | adult | | child | | adult | | child | | adult | |
| x->G | 563 | d->t | 419 | t | 820 | @ | 346 | n | 1324 | @ | 1126 |
| s->z | 352 | t->d | 396 | @ | 683 | n | 335 | h | 421 | n | 1041 |
| d->t | 334 | E->@ | 378 | s | 437 | t | 252 | @ | 172 | r | 686 |
| Y->@ | 333 | @->E | 324 | d | 411 | r | 219 | t | 159 | t | 519 |
| f->v | 332 | A->a | 255 | r | 281 | d | 219 | r | 124 | d | 405 |
| t->d | 187 | s->z | 215 | n | 261 | s | 107 | d | 74 | I | 267 |
| o->0: | 173 | I->E | 194 | z | 241 | h | 79 | G | 56 | s | 227 |
| z->s | 140 | E+->@ | 183 | A | 159 | G | 77 | s | 53 | h | 225 |
| m->n | 130 | z->s | 174 | j | 152 | E | 77 | v | 43 | E | 196 |
| E->@ | 127 | @->I | 172 | a | 116 | j | 72 | a | 41 | I | 188 |

Table 4: *ASR performance comparison at the word-level*

| Model | WER % (Child) | WER % (Adult) | Error Ratio |
|---|---|---|---|
| Wav2Vec2 Large | **13.2** | **11.1** [27] | **1.85** |
| XLSR Large | 17.6 | 15.7 [28] | 2.14 |
| MMS-NL | 21.0 | 14.5 [17] | 1.87 |
| Whisper | **9.8** | **6.7** [17] | **0.58** |

### 3.3. General Reading Error Detection and Classification

Table 5 presents the word-level reading error detection performance of the four ASR test models evaluated by precision, recall, and F1 scores. Overall, the results indicate that the best two performing models, Wav2Vec2 Large and Whisper each possess distinct strengths in reading error detection. On the one hand, Wav2Vec2 Large exhibits the highest performance. With a recall of 0.8, it can identify 80% of reading errors, although its precision is lower, especially for substitution (0.17) and deletion (0.47) errors, indicating a significant number of false alarms. While Whisper outperforms Wav2Vec2 in precision and detecting deletions, its recall of substitution (0.64) and insertion (0.17) is lower than the rest, leading to numerous misdetections.

### 3.4. Reading Miscue Detection

The two ASR models demonstrating the best performance in terms of WER (Section 3.2) and general error detection (Section 3.3), Wav2Vec2 Large and Whisper are selected for reading miscue detection and further analysis. Table 6 shows the detection performance of the reading miscue defined in Table 1 among two best models, evaluated by precision, recall, and F1 scores. Gray cells highlight scores of a miscue detection task where the model's performance is superior compared to similar tasks in error detection. Results show that Whisper performs better on precision (0.52) and F1 (0.52) for miscue detection, especially good at deletion (F1 at 0.71), orthographic (F1 at 0.36) and Non-semantic/orthographic substitution (F1 at 0.51), while Wav2Vec2 performs better on recall (0.83) of miscue detection and is particularly good at detecting insertion (F1 at 0.63) and semantic substitution (F1 at 0.39).

Table 7 shows the recognition accuracy of three different reading attempts for the two best error detection models. It displays that the correctly read word recognition accuracy is high for both models (0.90 and 0.93), while performance recognition on incorrectly read attempts is lower (0.60, 0.53) and considerably lower on part-of-word attempts (0.29 and 0.23).

Table 8 presents the different types and frequencies of false recognition cases made by the best two-performing models of this work when recognizing incorrectly read attempts. For Wav2Vec2 Large, the majority of false recognitions are substitution errors (211 cases), with varied error patterns including rectify incorrect words, recognize incorrect words into more than one word or merge them with subsequent sounds. We observe many word substitutions cases due to incorrectly decoding consonants (e.g., 'groot' to 'goot') and Wav2Vec2 tends to omit spelling/syllable sounds like 'a' or 'de'. Conversely, for Whisper, the majority of false recognitions (207 cases) is omitting incorrect attempts, with few substitution errors to single word.

## 4. Discussion

To address RQ1, namely, how effectively pretrained ASR models, initially trained on adult speech, recognize speech of native children in primary school, we analyze the results on two levels: phoneme and word. The results presented at the phoneme-level in Section 3.1 indicate that the phoneme recognition performance of current SOTA Dutch ASR models for children's speech is still notably imperfect, in line with the literature [8, 6]. In our experiment, the Hubert Large model outperforms other phoneme-based models when tested on both children's speech (Jasmin dataset, 23.1% PER) and adult speech using the Common Voice test dataset (6.3% PER). The primary misrecognized phonemes and specially difficult cases for children aligns with findings reported in [5], which suggests bias in ASR and imbalance in dataset. As illustrated in Table 3, the errors made by ASR models in recognizing Dutch-speaking children speech differ from those made in adult speech. These models tend to make phonological recognition errors in recognizing child speech that not frequently happened in adult speech, such as substituting /G/ by /x/, /@/ by /Y/, and /v/ by /f/. Furthermore, it is interesting to observe that while recognizing children speech, ASR frequently omits phonemes not only typical of adults, such as /@/, /n/, /t/, or /r/, but also specific cases unique to children, such as /z/, /A/, and /a/, while also unnecessarily inserting phonemes like /G/, /v/ or /a/.

On the contrary, as evidenced in Section 3.2, word-level Dutch children speech recognition yields more favorable performance outcomes compared to phoneme-level recognition, as reported in Table 4. whisper and Wav2Vec2 Large present the lowest WERs, when evaluated on children's speech (Jasmin dataset, 9.8% and 13.2%, respectively) and adult speech (Common Voice dataset, 6.7% and 11.7%, respectively). Wav2Vec2-family models achieve an Error Ratio higher than 1 and whisper;s operates conversely. Further analysis, see Table 8, shows that this is because Wav2Vec2 Large tends to produce more errors while Whisper tends to ignores errors. This suggests that Error Ratio can be an index to identify different decoding strategies for incorrect speech.

To address RQ2, in which we assess the capability of current SOTA ASR models in detecting children's reading miscues, we analyzed machine learning metrics including precision, recall, and F1 scores for miscue detection. In the first step, while focusing specifically on broad error categories including insertions, deletions, and substitutions, we found that Whisper achieves the highest precision, albeit with room for improvement (0.54, see Table 5). Notably, this model excels in detecting deletion errors with precision, recall, and F1 scores of 0.60, 0.85, and 0.71, respectively. Conversely, Wav2Vec2 Large exhibits the best overall recall (0.80) and F1 score (0.56), indicating its effectiveness in identifying actual error instances, albeit potentially at the expense of precision. Specifically, this model performs exceptionally well in classifying insertion errors with recall and F1 scores of 0.80 and 0.77, respectively.

Table 5: *Error classification at word-level with loose location and error type*

| Model | All errors | | | Insertion (72.9%) | | | Substitution (18.7%) | | | Deletion (8.4%) | | |
|---|---|---|---|---|---|---|---|---|---|---|---|---|
| | Precision | Recall | F1 | Precision | Recall | F1 | Precision | Recall | F1 | Precision | Recall | F1 |
| Wav2Vec2 Large | 0.43 | **0.80** | **0.56** | 0.74 | **0.80** | **0.77** | 0.17 | 0.81 | 0.28 | 0.47 | 0.79 | 0.59 |
| XLSR Large | 0.33 | 0.70 | 0.45 | 0.71 | 0.67 | 0.69 | 0.13 | **0.82** | 0.22 | 0.24 | 0.73 | 0.36 |
| MMS-NL | 0.17 | 0.32 | 0.22 | 0.70 | 0.17 | 0.26 | 0.11 | 0.71 | 0.19 | 0.14 | 0.75 | 0.24 |
| Whisper | **0.54** | 0.31 | 0.39 | **0.83** | 0.16 | 0.27 | **0.37** | 0.64 | **0.48** | **0.60** | **0.85** | **0.71** |

Table 6: *Detection of reading miscue in Table 1 by Wav2Vec2 Large and Whisper (P:Precision, R:Recall)*

| Model | All miscues | | | $I_m$ | D | OS | SS | O |
|---|---|---|---|---|---|---|---|---|
| | P | R | F1 | F1 | F1 | F1 | F1 | F1 |
| Wav2Vec2 | 0.29 | **0.83** | 0.43 | **0.63** | 0.59 | 0.17 | **0.39** | 0.28 |
| Whisper | **0.52** | 0.53 | **0.52** | 0.46 | **0.71** | **0.36** | 0.33 | **0.51** |

Table 7: *Accuracy for correctly/incorrectly read speech*

| Model | Correctly read | | Incorrectly read (586) |
|---|---|---|---|
| | Word (11322) | Part-of-word (180) | |
| Wav2Vec2 Large | 0.90 | **0.29** | **0.60** |
| Whisper | **0.93** | 0.23 | 0.53 |

Table 8: *Types and occurrence counts of false recognition made by Wav2Vec2 Large and Whisper for incorrectly read speech*

| Type of ASR false recognition | Wav2Vec2 | Whisper |
|---|---|---|
| ASR omits incorrectly read attempts | 22 | 202 |
| Rectify incorrect to correct word | 51 | 24 |
| Replace with a single word | 112 | 41 |
| Replace with 2 or more words | 29 | 0 |
| Merge incorrect sounds and subsequent pronunciation | 17 | 0 |
| Errors in manual transcription | 2 | 2 |

The advantages of Whisper and Wav2Vec2 Large are similarly evident in reading miscue detection. Compared miscue detection in Table 6 to error detection in Table 5, for Wav2Vec2 Large, while recall is slightly higher (0.83 for miscue vs 0.80 for errors), precision (0.29 for miscue vs 0.43 for errors) and performance on insertion miscues (F1 at 0.63 for miscue vs F1 at 0.77 for errors) are notably worse. Considering the removal of restart in insertion miscues, this suggests that Wav2Vec2 Large cannot support effectively differentiating restarts from other insertion errors. Conversely, for Whisper, recall and F1 scores for miscue detection are 0.53 and 0.52, compared to 0.31 and 0.39 in error detection. Insertion miscue detection of whisper is notably higher (F1 at 0.46) compared to that in error detection (F1 at 0.27). This suggests that Whisper has an advantage in eliminating restart errors.

To gain further insights into the ASR models' performance regarding accurately and inaccurately read words and part-of-words, we observe in Table 7 that the top-performing models, Wav2Vec2 Large and Whisper, manage to detect correctly read words (0.9 and 0.93 accuracy, respectively), but are less effective in identifying incorrectly read words (0.6 and 0.53 accuracy, respectively). Notably, both models perform inadequately when it comes to part-of-words. Wav2Vec2 Large is slightly better than Whisper on part-of-word and incorrectly read attempts recognition, but Whisper is more accurate on correct word recognition. This suggests differing decoding strategies for Whisper and Wav2Vec2 Large models and best WER may not indicate best reading miscue detection.

Finally, in order to understand the results of the error classification of incorrectly read words returned by the two top-performing models, Wav2Vec2 Large and Whisper, we offer more insights in Table 8. Whisper in 73% of the cases tries to just ignore the incorrectly read word, instead of providing semantically similar words, partial words, hesitations, combined words, or broken words. On the other hand, Wav2Vec2 Large offers a more varied output such as providing another word (48% of the cases), merge text (7% of the cases), or even splitting the words into two or more words (12%). This evidence leads us to consider Wav2Vec2 Large as a better system for recognizing incorrectly read speech.

One of the limitation of this work is that reading miscue ground truth labels are generated by automatic similarity scores. Possible discrepancies with human categorizations of reading miscues and alignment errors are not fully explored. These aspects will be addressed in future extensions of our work.

## 5. Conclusions

This paper investigated the potential of different SOTA pre-trained ASR models for enhancing word and phoneme reading diagnosis in Dutch native children. Results demonstrated significant contributions, with Hubert Large achieving an acceptable PER rate of 23.1%, while Whisper achieved a low WER of 9.8%. Although proficient, these ASR models still require refinement through the incorporation of domain-specific data to achieve greater precision and efficacy in reading miscue detection. Our findings also highlight the notable performance of Wav2Vec2 Large, which emerges as a frontrunner with a recall of 0.83 for automatic reading miscue detection, whereas Whisper exhibits the highest precision at 0.52 and an F1 score of 0.52. These advancements are encouraging, but we still need to harness and improve ASR technology for precise reading diagnosis. However, although current systems may not yet be ready for precise reading miscue classification like human experts, they can still be valuable in reading education as assistant tools. For instance, they can aid in the initial screening of general reading difficulties before a comprehensive diagnosis, ultimately alleviating the workload for teachers.

Future research may address finetuning ASR in children reading speech and resolving data bias in differentiating words with consonant confusion pairs, as suggested in our findings. This could be done, considering the frequent emergence of SOTA ASR models, following a data-centric approach [31], focusing mainly on the data of the models to gain better insights.

## 6. Acknowledgements

This work was supported by the NWO research programme AiNed Fellowship Grants under the project Responsible AI for Voice Diagnostics (RAIVD) - NGF.1607.22.013.